\documentclass[conference]{IEEEtran}
\IEEEoverridecommandlockouts
\usepackage{amsmath,amssymb,amsfonts}
\usepackage{algorithmic}
\usepackage{graphicx}
\usepackage{textcomp}
\usepackage{xcolor}
\def\BibTeX{{\rm B\kern-.05em{\sc i\kern-.025em b}\kern-.08em
    T\kern-.1667em\lower.7ex\hbox{E}\kern-.125emX}}

\usepackage{amsmath,amsfonts,bm}
\usepackage{amsthm}
\usepackage{amssymb}
\usepackage{mathrsfs}

\usepackage{graphicx}
\usepackage{subfigure}

\newtheorem{thm}{Theorem}
\newtheorem{rmk}{Remark}

\newtheorem{corollary}{Corollary}

\newtheorem{assumption}{Assumption}

\def\bfa{{\boldsymbol a}}

\def\bfo{{\boldsymbol o}}
\def\bfp{{\boldsymbol p}}
\def\bfq{{\boldsymbol q}}
\def\bfr{{\boldsymbol r}}

\def\bfx{{\boldsymbol x}}

\def\bfmu{{\boldsymbol \mu}}

\def\bfA{{\boldsymbol A}}

\def\bfI{{\boldsymbol I}}

\def\bfW{{\boldsymbol W}}
\def\bfX{{\boldsymbol X}}

\newcommand{\be}{\begin{equation}}
\newcommand{\ee}{\end{equation}}

\def\eqref#1{equation~\ref{#1}}

\def\1{\bm{1}}

\DeclareMathAlphabet{\mathsfit}{\encodingdefault}{\sfdefault}{m}{sl}
\SetMathAlphabet{\mathsfit}{bold}{\encodingdefault}{\sfdefault}{bx}{n}

\usepackage{tikz}
\usepackage{textcomp}
\usepackage{hyperref}
\usepackage{lipsum}
\newcommand\copyrighttext{%
  \footnotesize \textcopyright 2024 IEEE. Personal use of this material is permitted.
  Permission from IEEE must be obtained for all other uses, in any current or future
  media, including reprinting/republishing this material for advertising or promotional
  purposes, creating new collective works, for resale or redistribution to servers or
  lists, or reuse of any copyrighted component of this work in other works.
  }
\newcommand\copyrightnotice{%
\begin{tikzpicture}[remember picture,overlay]
\node[anchor=south,yshift=10pt] at (current page.south) {\fbox{\parbox{\dimexpr\textwidth-\fboxsep-\fboxrule\relax}{\copyrighttext}}};
\end{tikzpicture}%
}

\begin{document}

\title{Learning on Transformers is Provable Low-Rank and Sparse: A One-layer Analysis\\
\thanks{This work is supported by Rensselaer-IBM AI Research Collaboration (http://airc.rpi.edu), part of the IBM AI Horizons Network (http://ibm.biz/AIHorizons).}
}

\author{\IEEEauthorblockN{Hongkang Li}
\IEEEauthorblockA{\textit{Dept. Electrical, Computer, and System Engineering} \\
\textit{Rensselaer Polytechnic Institute}\\
Troy, NY, USA \\
lih35@rpi.edu}
\and
\IEEEauthorblockN{Meng Wang}
\IEEEauthorblockA{\textit{Dept. Electrical, Computer, and System Engineering} \\
\textit{Rensselaer Polytechnic Institute}\\
Troy, NY, USA\\
wangm7@rpi.edu}
\and
\IEEEauthorblockN{Shuai Zhang}
\IEEEauthorblockA{\textit{Dept. Data Science} \\
\textit{New Jersey Institute of Technology}\\
Newark, NJ, USA\\
sz457@njit.edu}
\and
\IEEEauthorblockN{Sijia Liu}
\IEEEauthorblockA{\textit{Dept. Computer Science and Engineering} \\
\textit{Michigan State University}\\
East Lansing, MI, USA\\
liusiji5@msu.edu}
\and
\IEEEauthorblockN{Pin-Yu Chen}
\IEEEauthorblockA{\textit{IBM Thomas J. Watson Research Center} \\
Yorktown Heights, NY, USA\\
pin-yu.chen@ibm.com}
}

\maketitle
\copyrightnotice
\begin{abstract}
Efficient training and inference algorithms, such as low-rank adaption and model pruning, have shown impressive performance for learning Transformer-based large foundation models. However, due to the technical challenges of the non-convex optimization caused by the complicated architecture of Transformers, the theoretical study of why these methods can be applied to learn Transformers is mostly elusive. To the best of our knowledge, this paper shows the first theoretical analysis of the property of low-rank and sparsity of one-layer Transformers by characterizing the trained model after convergence using stochastic gradient descent. By focusing on a data model based on label-relevant and label-irrelevant patterns, we quantify that the gradient updates of trainable parameters are low-rank, which depends on the number of label-relevant patterns. We also analyze how model pruning affects the generalization while improving computation efficiency and conclude that proper magnitude-based pruning has a slight effect on the testing performance. We implement numerical experiments to support our findings. 
\end{abstract}

\begin{IEEEkeywords}
Transformer, low-rank adaption, model pruning, mechanism
\end{IEEEkeywords}

\section{Introduction}

Transformers \cite{VSPU17} have recently achieved remarkable empirical success in many areas, including natural language processing \cite{ BMRS20, BCEG23}, computer vision \cite{DBKW20, LLCH21}, and reinforcement learning \cite{CLRL21, OWJA22}. Due to the more complicated architecture and the larger size of transformer parameters, numerous modern fine-tuning methods are proposed to adapt to downstream tasks efficiently and effectively. For example, Low-Rank Adaptation (LoRA) \cite{HWAL21, DPHZ23} fixes the pre-trained model and only tunes weights in the newly added low-rank modules, which leads to low-rank gradient updates during fine-tuning. Model pruning \cite{HMD15, FA23, SLBK23} compresses the model by removing unimportant model weights to reduce the computation cost. Despite immense success in experiments for efficient learning, one critical question remains less explored, which is 
\begin{center}
\textit{Why do LoRA and model pruning work for Transformer learning and generalization?}
\end{center}
Some existing theoretical works \cite{BBSS22, AAM22, DLS22, AAM23} probe the rank changes in neural network training and show that the low-rank bias exists if the labeling function is determined by a low-dimensional subspace of the input. Most of these works focus on two-layer non-transformer networks. Only \cite{BLAB23} is established for Transformers, which is based on training diagonal weight parameters and concludes that the difference between trained and initial weights has a low rank. Other existing theoretical works \cite{YW23, YLGW23, ZWCL23} consider the generalization of model pruning. However, these works are built on convolutional neural networks, and no theoretical works are for Transformer-based models.

\subsection{Major Contributions}

Our work studies the mechanism of low-rank property and model pruning (i) by characterizing the directions and magnitudes of the updates of the trained Transformer (ii) without assuming diagonal matrices for training. The theoretical analysis is built upon a data model where tokens are noisy versions of two \textit{label-relevant} patterns that determine the binary label and \textit{label-irrelevant} patterns that do not decide the label as \cite{LWLC23}. The contributions are as follows. 

\textbf{First, we quantitatively characterize the trained parameters, which imply a property of low rank.} We prove that the differences between trained weights and initialization in Transformers map patterns to directions of either one of the label-relevant patterns or a certain linear positive combination of label-relevant patterns with a large magnitude. This indicates that the gradient updates in the trained model have a low-rank property, which corresponds to the number of label-relevant patterns.

\textbf{Second, we show that magnitude-based pruning preserves the generalization performance.} We quantify the generalization if neurons with the smallest magnitude after training in $\bfW_O$ are removed and prove that the generalization error is almost the same even when a constant fraction of neurons are removed. Conversely, the generalization error is shown to be at least $\Omega(R)$ with $R$ fraction of neurons with large magnitude removed.   

\textbf{Third, we verify our findings of low rank and model pruning with experiments. } Experimental results show that the reconstructed gradient updates via low-rank approximation enable a similar test performance if the rank is no smaller than the number of label-relevant patterns. We also justify the difference between pruning neurons with small magnitude or large magnitude.

\subsection{Related Works}

\textbf{The optimization and generalization of Transformers} There are several works about the optimization and generalization analysis of fine-tuning or prompt tuning of Transformers. \cite{JSL22, LWLC23, ZLYC24} explore the generalization performance of one-layer Transformers with sample complexity by assuming spatial association or the majority voting of tokens. \cite{LWML23, L23} investigate the effect of the relative positional encoding in training for Graph Transformer.
\cite{LLR23} study how one-layer Transformers learn semantic structure along the training. \cite{ORST23} characterize the trajectory of the gradient flow of prompt tuning in training attention networks. \cite{TLZO23, TLTO23} show that prompt tuning or fine-tuning on Transformers converge to a max-margin SVM solution.   \cite{TWCD23, TWZC23} explore the training dynamics of Transformers for next token prediction when the length of sequences is infinite. \cite{HCL23, LWLW23, LWLC24} study the training and generalization of Transformers for in-context learning.

\textbf{Neural network learning on structured data.} Some existing works study the learning and generalization of CNN-type neural networks, where the data distribution follows certain specific structures. 
\cite{LL18} consider learning on neural networks with data from 
separated distributions. \cite{LWLC22, SLW24} study the generalization on graph-structured data. \cite{LZW22, LZZW24} characterize the loss landscape of a one-layer fully-connected neural network when the data distribution follows a mixture of Gaussians that leads to a group imbalance in data. By assuming discriminative patterns and background patterns in data, \cite{SWL21, KWLS21, ZWCL23} delve into the generalization analysis of fully connected networks and convolutional neural networks. \cite{WL21} adapt such analysis to the area of self-supervised contrastive learning. All these works consider neural networks without self-attention.
\section{Problem Formulation}
We study a 
binary classification problem given $N$ 
 training samples $\{(\bfX^n, y^n)\}_{n=1}^N$ generated from a distribution $\mathcal{D}$. Each data point $\bfX^n$ contains $L$ tokens $\bfx_1^n,\ \bfx_2^n,\cdots,\bfx_L^n$, i.e.,  $\bfX^n=[\bfx_1^n,\cdots,\bfx_L^n]\in\mathbb{R}^{d\times L}$, where each token is $d$-dimensional and unit-norm. $y^n\in\{+1,-1\}$ is a scalar. 

Learning is implemented with a one-layer Transformer, i.e., a neural network with a single-head self-attention layer and a two-layer fully connected network, as shown in (\ref{eqn: attention}). Let  $\mathcal{S}^n \subseteq [L]$ denote the set of indices of tokens in $\bfX^n$ used in the output computation. This is simplified from practical Transformers \cite{VSPU17, DBKW20} as considered in \cite{JSL22, LWLC23, ORST23}. 

\begin{equation}
\begin{aligned}
    F(\bfX^n)=&\frac{1}{|\mathcal{S}^n|}\sum_{l\in\mathcal{S}^n}\bfa_{(l)}^\top\text{Relu}(\bfW_O\bfW_V\bfX^n\\
    &\cdot\text{softmax}({\bfX^n}^\top\bfW_K^\top\bfW_Q\bfx_l^n)),
\end{aligned}\label{eqn: attention}
\end{equation}
where the queue weights   $\bfW_Q\in\mathbb{R}^{m_b\times d}$, the key weights  $\bfW_K\in\mathbb{R}^{m_b\times d}$, and the value weights  $\bfW_V\in\mathbb{R}^{m_a\times d}$. 
$\bfW_O\in\mathbb{R}^{m\times m_a}$ and $\bfA=(\bfa_{(1)},\bfa_{(2)},\cdots,\bfa_{L})$ where $\bfa_{(l)}\in\mathbb{R}^{m},\ l\in[L]$ are the hidden- and output-layer weights of the two-layer perceptron, respectively. $m$ is the number of neurons in the hidden layer. $\text{Relu}: \mathbb{R}^{m}\rightarrow \mathbb{R}^m$ where $\text{Relu}(\bfx)=\max\{\bfx,0\}$. $\text{softmax}: \mathbb{R}^{L}\rightarrow \mathbb{R}^L$ where $\text{softmax}(\bfx)=(e^{x_1}, e^{x_2}, \cdots, e^{x_L})/\sum_{i=1}^L e^{x_i}$. Let $\Psi=\{\bfA,\bfW_O,\bfW_V,\bfW_K,\bfW_Q\}$ be the set of trainable parameters.  The training problem minimizes the empirical risk $f_N(\Psi)$, 
\begin{equation}
    \min_{\psi}: f_N(\psi)=\frac{1}{N}\sum_{n=1}^N \ell(\bfX^n,y^n;\psi), \label{eqn: min_train}
\end{equation}
where $\ell(\bfX^n,y^n;\Psi)$ is the Hinge loss function, i.e.,
\begin{equation}
    \ell(\bfX^n,y^n;\Psi)=\max\{1-y^n\cdot F(\bfX^n),0\}.
\end{equation}

The generalization performance of a learned model $\Psi$ 
is evaluated by the population risk $f(\Psi)$, where 
\begin{equation}
    f(\Psi)=\mathbb{E}_{(\bfX,y)\sim\mathcal{D}}[\ell(\bfX,y;\Psi)].
\end{equation}

The training problem (\ref{eqn: min_train}) is solved by a stochastic gradient descent (SGD),  
$\bfW_Q$, $\bfW_K$ and $\bfW_V$ are initialized such that all diagonal entries are set as $\delta$ with $\delta\in(0,0.2]$, and all other entries are $0$.
Every entry of $\bfW_{O}$ is generated from  $\mathcal{N}(0,\xi^2)$. 
Every entry of $\bfa_{l}^{(0)}$ is sampled from $\{+\frac{1}{\sqrt{m}},-\frac{1}{\sqrt{m}}\}$ with equal probability. $\bfA$ does not update during the training. At iteration $t$, $t=0,1,2,\cdots,T-1$, the gradient is computed using a batch $\mathcal{B}_t$ with $|\mathcal{B}_t|=B$. The step size is $\eta$. One step of gradient update at iteration $t$ is that for any $\bfW\in\Psi\backslash\bfA$, 
\begin{equation}\label{eqn:gradient}
\begin{aligned}
\bfW^{(t+1)}&=\bfW^{(t)}-\eta \cdot \frac{1}{B} \sum_{n\in\mathcal{B}_t} \nabla_{\bfW^{(t)}}\ell(\bfX^n,y^n;\Psi).
\end{aligned}
\end{equation}

\section{Theoretical Results}


This section is organized as follows. We first formulate the data model with patterns we use for analysis in Section \ref{subsec: data_model}. Then, we introduce the main theoretical results on the low-rank property of gradient updates and the sparsity of the trained model associated with model pruning in Section \ref{subsec: theory}.

\subsection{Data Model}\label{subsec: data_model}

Consider that there are  $M\ (\Theta(1)<M<m_a,m_b=\Theta(M))$ distinct patterns $\mathcal{U}=\{\bfmu_1,\ \bfmu_2,\cdots, \bfmu_M\}$ in $\mathbb{R}^d$ that are pairwise orthogonal with unit norm as in \cite{ORST23, HCL23, AL23}. $\bfmu_1,\bfmu_2$ are \textit{discriminative patterns} that determine the binary labels, 
and the remaining $M-2$ patterns $\bfmu_3,\ \bfmu_4,\cdots,\bfmu_M$ are \textit{non-discriminative patterns} that do not affect the labels.   
Each token $\bfx_l^n$ of $\bfX^n$ is a noisy version of one of the patterns, i.e., 
\begin{equation}\min_{j\in[M]}\|\bfx_l^n-\bfmu_j\|\leq  \tau, 
\label{eqn: mu_j}\end{equation}
and the noise level $\tau < O(1/M)$. 

Following \cite{LWLC23, LWML23}, the label $y^n$  is determined by the tokens that correspond to discriminative patterns via a majority vote. Specifically, if the number of tokens that are noisy versions of $\bfmu_1$ (or $\bfmu_2$) is larger than the number of tokens of $\bfmu_2$ (or $\bfmu_1$) in $\bfX^n$, then $y^n=1$ (or $y^n=-1$). In the case that the label $y^n=1$ (or $y^n=-1$), the tokens that are noisy $\bfmu_1$ (or $\bfmu_2$) are called \textit{label-relevant} tokens, and the tokens that are noisy $\bfmu_2$ (or $\bfmu_1$) are named \textit{confusion} tokens. 
All other tokens that correspond to $\bfmu_i$, $3\leq i\leq M$ are called label-irrelevant tokens.  

We assume a balanced dataset. Let
$\mathcal{D}_+=\{(\bfX^n,y^n)|y^n=+1, n\in[N]\}$ and $\mathcal{D}_-=\{(\bfX^n,y^n)|y^n=-1, n\in[N]\}$ be the sets of positive and negative labels, respectively. Then 
$\Big||\mathcal{D}^+|-|\mathcal{D}^-|\Big|=O(\sqrt{N})$.


\subsection{Formal Theoretical Results}\label{subsec: theory}

\begin{table}[h!]
  \begin{center}
        \caption{Some important notations}    \label{tab: notation}
    \begin{tabular}{l|c} 
 \hline \small$\delta$ & \scriptsize{The magnitude of the initialization of query,  key, and value matrices}\\
 \hline
 \small$\tau$ & \scriptsize{Token noise level}\\
 \hline
  \small$M$ & \scriptsize{Total number of patterns}\\
  \hline
  \small$m$ & \scriptsize{The number of neurons in $\bfW_O$}\\
 \hline

    \end{tabular}

  \end{center}
\end{table}
Some key notations are summarized as in Table \ref{tab: notation}. Then, we introduce the major theoretical results in the following.

\begin{thm} (low rank and sparsity) \label{thm: rank}
Suppose all conditions for a zero generalization $f(\Psi)=0$ in Theorem 1 in \cite{LWLC23} holds. Denote $\Delta\bfW=\bfW^{(T)}-\bfW^{(0)}$ as the gradient update of $\bfW$ after convergence. Let $\alpha_*$ be the average fraction of label-relevant tokens in each data.  


Then, the trained model after $T=\Theta(\eta^{-3/5}\alpha_*^{-1})$ iterations with a batch size of $B\geq\Omega(M)$ satisfies that 
\begin{enumerate}
\item if $\bfW\in\{\bfW_Q, \bfW_K\}$, , then for any $\bfmu\in\{\bfmu_1,\bfmu_2\}$, and $\bfmu'\in\mathcal{U}\backslash\{\bfmu\}$, we have
\begin{equation}
    {\bfmu}^\top\Delta\bfW\bfmu\geq \Omega(\sqrt{\log T}),\ {\bfmu'}^\top\Delta\bfW\bfmu\leq O(1/(MT)),\label{eqn: QK1}
\end{equation}
and for any $\bfmu\in\mathcal{U}\backslash\{\bfmu_1,\bfmu_2\}$, $\bfmu'\in\{\bfmu_1,\bfmu_2\}$, and $\bfmu''\in\mathcal{U}\backslash\{\bfmu\}$, we have
\begin{equation}
\begin{aligned}
    &{\bfmu'}^\top\Delta\bfW\bfmu\geq \Omega(1),\ {\bfmu}^\top\Delta\bfW\bfmu\leq O(\delta),\\
    &{\bfmu''}^\top\Delta\bfW\bfmu\leq O(1/M)\label{eqn: QK2}
\end{aligned}
\end{equation}

\item For $\bfmu\in\{\bfmu_1,\bfmu_2\}$, $\bfmu'\in\mathcal{U}\backslash\{\bfmu\}$, we have
\begin{equation}
    \bfmu^\top\Delta\bfW_V\bfmu\geq \Omega(1), {\bfmu'}^\top\Delta\bfW_V\bfmu\leq O(1/(MT)),\label{eqn: V1}
\end{equation}
for $\bfmu\in\mathcal{U}\backslash\{\bfmu_1,\bfmu_2\}$, $\bfmu'\in\mathcal{U}\backslash\{\bfmu\}$, we have
\begin{equation}
    \bfmu^\top\Delta\bfW_V\bfmu\geq \Omega(1/M), {\bfmu'}^\top\Delta\bfW_V\bfmu\leq O(1/M),\label{eqn: V2}
\end{equation}
\item Let $\bfo_i$ denote the $i$-th row of $\bfW_{O}$. There exists a set $\mathcal{L}\subset[m]$ with the size $|\mathcal{L}|=\Theta(m)$ such that for $i$ with $a_i>0$ and $\bfo_i'\in\{\bfo_i,\Delta\bfo_i\}$, 
\begin{equation}
\begin{aligned}
    \|{\bfo_i'}^{(T)}\bfmu_1\|\geq \Omega(1/\sqrt{m}),\  &i\in\mathcal{L}\\
    \|{\bfo_i'}^{(T)}\bfmu\|\leq O(1/(\sqrt{m})),\  &i\notin\mathcal{L}, \bfmu\in\mathcal{U}\backslash\{\bfmu_1\},
    \end{aligned}\label{eqn: L ai>0}
\end{equation}
for $i$ such that $a_i<0$,
\begin{equation}
\begin{aligned}
    \|{\bfo_i'}^{(T)}\bfmu_2\|\geq \Omega(1/\sqrt{m}),\  &i\in\mathcal{L}\\
    \|{\bfo_i'}^{(T)}\bfmu\|\leq O(1/(\sqrt{m})),\  &i\notin\mathcal{L}, \bfmu\neq\mathcal{U}\backslash\{\bfmu_2\},
    \end{aligned}\label{eqn: L ai<0}
\end{equation}

and for $i\notin\mathcal{L}$, 
\begin{equation}
    \|{\bfo_i'}^{(T)}\|\geq \Omega(1/(\sqrt{m}))\label{eqn: small WO}
\end{equation}
\end{enumerate}

\end{thm}

Theorem \ref{thm: rank} characterizes the directions of the gradient updates of all the trainable matrices onto $\bfmu\in\mathcal{U}$. Specifically, $\Delta\bfW_Q$, $\Delta\bfW_K$ map all directions in $\mathcal{U}$ to $\bfmu_1$, $\bfmu_2$, or a positive linear combination of $\bfmu_1$ and $\bfmu_2$, with the projection onto $\bfmu_1$ or $\bfmu_2$ order-wise the largest to be $\Omega(\sqrt{\log T})$ by (\ref{eqn: QK1}) and (\ref{eqn: QK2}). The result is the same for $\Delta\bfW_V$ by (\ref{eqn: V1}) and (\ref{eqn: V2}), but with the largest projection $\Omega(1)$. The trained $\bfW_O$ has a property that, except for a small fraction of neurons with a small magnitude, all other neurons and their gradient updates are close to either $\bfmu_1$ or $\bfmu_2$ with a larger magnitude from (\ref{eqn: L ai>0}), (\ref{eqn: L ai<0}), and (\ref{eqn: small WO}). 

\begin{rmk}\label{rmk: rank}
Theorem \ref{thm: rank} implies a low-rank property of the gradient updates through learning since all gradient updates map unit vectors in $\mathbb{R}^M$ close to either $\bfmu_1$, $\bfmu_2$, or a certain positive linear combination of $\bfmu_1$ and $\bfmu_2$. We can infer that $\Delta\bfW_Q$, $\Delta\bfW_K$, $\Delta\bfW_V$, and $\Delta\bfW_O$ can be approximated by rank $2$, which is the number of label-relevant patterns that determine the complexity of the classification task. Meanwhile, the results in (\ref{eqn: L ai>0}), (\ref{eqn: L ai<0}), and (\ref{eqn: small WO}) show that a constant fraction of neurons of $\bfW_O$ has a much small magnitude compared with remaining ones, indicating the sparsity of the trained $\bfW_O$. 
\end{rmk}

The proof of Theorem \ref{thm: rank} can be established by further deriving the conclusions in Claim 2 of \cite{LWLC23} given the convergence condition. Theorem \ref{thm: rank} leads to the following corollary on pruning. 

\begin{corollary} (Model Pruning)\label{cor: pruning}
Pruning all neurons $i\in\mathcal{L}^c$ of $\bfW_O$ leads to a generalization error  
    \vspace{-2mm}
\begin{equation} f(\Psi_{\mathcal{L}^c})\leq {O}(1/M),
    \end{equation}
    where $\Psi_{\mathcal{L}^c}$ represents the model weights after removing neurons in $\mathcal{L}^c$ in $\bfW_O$.
   In contrast, pruning $\mathcal{S} \subset \mathcal{L}$ with size $|\mathcal{S}|=Rm$, where $R\in(0,1)$ and is a constant
   results in a generalization error of 
        \begin{equation}
        f(\Psi_{\mathcal{S}})\geq \Omega( R).
    \end{equation}
\end{corollary}

Corollary \ref{cor: pruning} shows that magnitude-based pruning of $\bfW_O$ does not harm the generalization in our setup. Our results show that pruning neurons with a smaller magnitude slightly increases the generalization error compared with that of the unpruned $\Psi$. Nevertheless, pruning neurons with a larger magnitude leads to a generalization error increasing as the pruning ratio $R$ increases. Corollary \ref{cor: pruning} indicates that in our setup, magnitude-based pruning on $\bfW_O$ does not hurt the model's ICL capability.
\section{Numerical Experiments}

We present experiments using a one-layer Transformer (\ref{eqn: attention}) using synthetic data. Set the dimension of data and embeddings to be $d=m_a=m_b=20$. $M=20$. Let $\sigma=0.1$. Following \cite{LWLC23}, we generate each token of $\bfmu_i$ from a Gaussian distribution $\mathcal{N}(\bfmu_i, \sigma^2\cdot I)$ with the mean $\bfmu_i$ and covariance $\sigma^2\bfI$, where $\bfI\in\mathbb{R}^d$ is the identity matrix.
$\bfW_Q^{(0)}=\bfW_K^{(0)}=\bfW_V^{(0)}=\delta^2\bfI$ where $\delta=0.1$, and each entry of $\bfW_{O}^{(0)}$ follows $\mathcal{N}(0,\xi^2)$, $\xi=0.1$.  The number of neurons $m$ of $\bfW_O$ is  $200$. 
For simplicity, we set the fraction of different patterns the same among all the data.

\begin{figure}[!h]
\vspace*{-2mm}
\centering
\centerline{
\begin{tabular}{cc}
\includegraphics[width=.22\textwidth,height=1.65in]{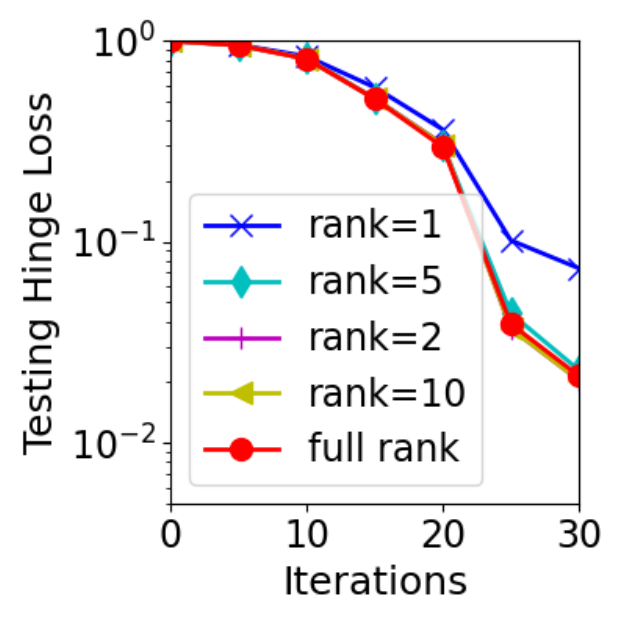}  
&
\hspace*{-1mm}
\includegraphics[width=.22\textwidth,height=1.65in]{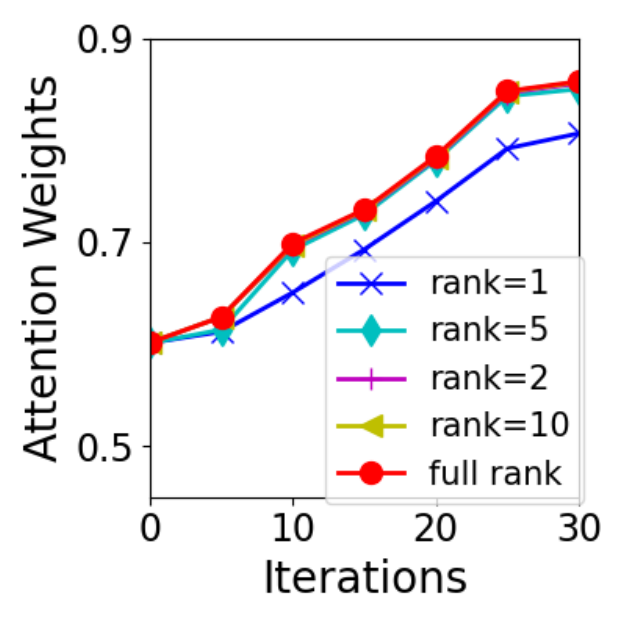}\vspace*{-1mm}
\\
(A) 
& \hspace*{-2mm} (B)
\end{tabular}}
\vspace*{-1mm}
\caption{When all trainable parameters are approximated with the rank equal to $1$, $2$, $5$, $10$, and $20$ (full rank), the results of (A) testing hinge loss, (B) the attention weights summation on label-relevant tokens. }\label{fig: rank loss}
\vspace{-1mm}
\end{figure}

We use different numbers of the rank to approximate the gradient updates of trainable parameters. Figure \ref{fig: rank loss} shows that when the rank is not smaller than $2$, which is the number of label-relevant patterns, the testing performance and the attention map are close to those with full-rank approximation. When the rank equals $1$, there is an evident drop in the testing performance and attention weights on label-relevant patterns. This is consistent with the disucssion of Remark \ref{rmk: rank}.

\begin{figure}[!h]
\vspace*{-2mm}
\centering
\centerline{
\begin{tabular}{cc}
\includegraphics[width=.17\textwidth,height=1.3in]{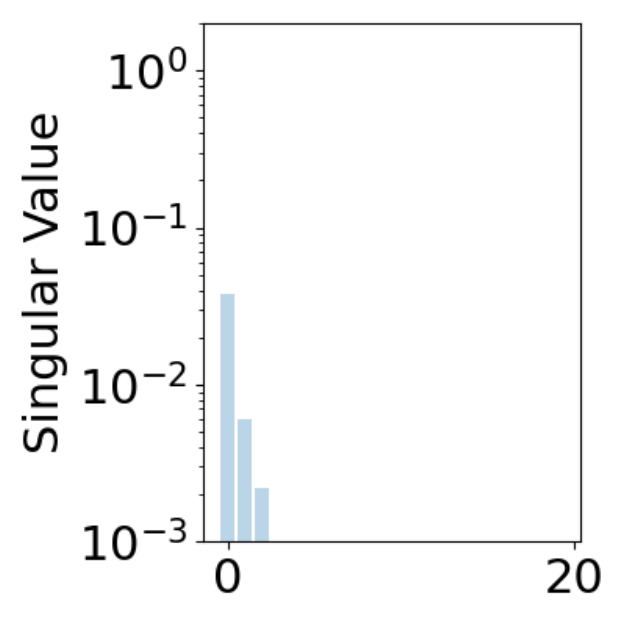}  
&
\hspace*{-1mm}
\includegraphics[width=.17\textwidth,height=1.3in]{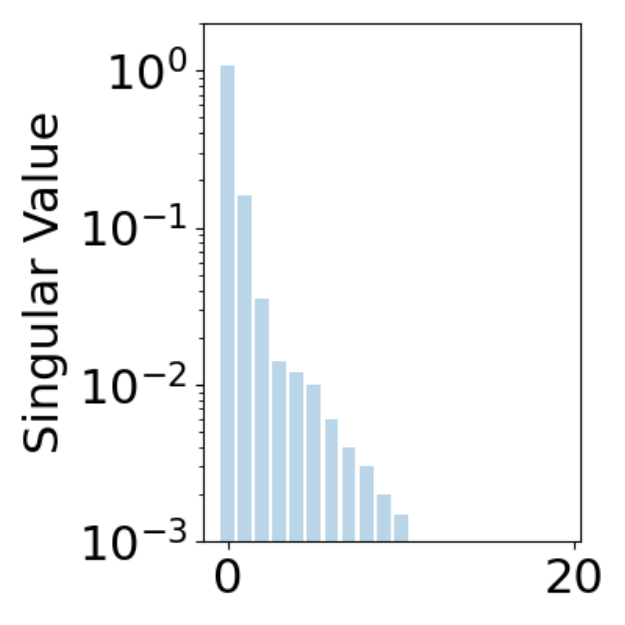}\vspace*{-1mm}\\
\hspace*{5mm}(A) 
& \hspace*{5mm} (B)\\
\includegraphics[width=.17\textwidth,height=1.3in]{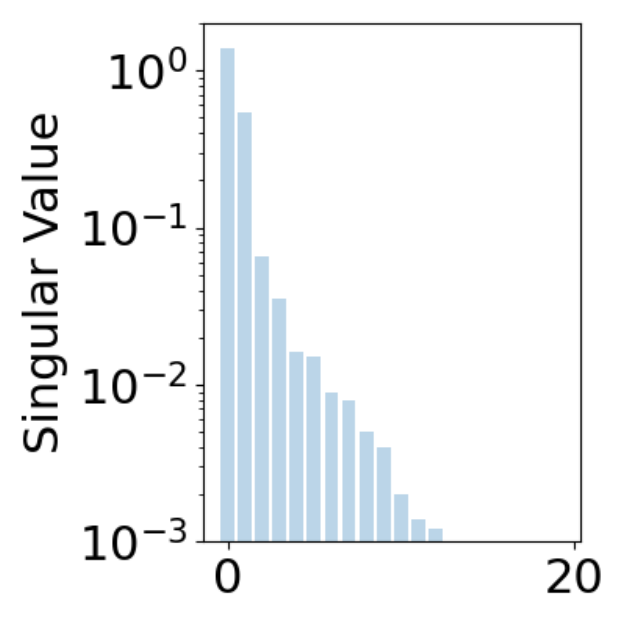}  
&
\hspace*{-1mm}
\includegraphics[width=.2\textwidth,height=1.3in]{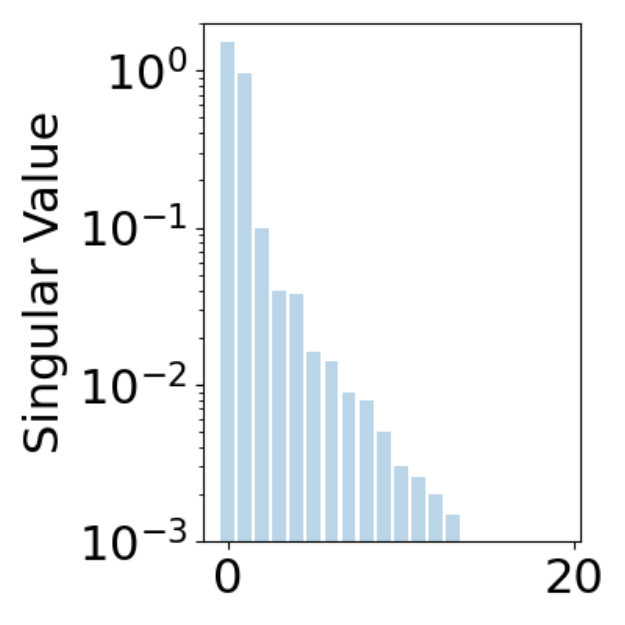}\vspace*{-1mm}
\\
\hspace*{+5mm}(C)
& \hspace*{+5mm} (D)\\
\end{tabular}}
\vspace*{-1mm}
\caption{The singular value of $\Delta\bfW_K^{(t)}$ when (A) $t=1$, (B) $t=10$, (C) $t=20$, (D) $t=30$. }\label{fig: rank WK}
\vspace{-1mm}
\end{figure}

We then observe the changes in the rank of parameter updates. We take $\bfW_K$ as an example. Figure \ref{fig: rank WK} shows that (1) the singular values of $\Delta\bfW_K^{(t)}$ increase along the training; (2) The first two singular values become the largest along the training, while others are at least $10$ times smaller. This means that the ranks of parameter updates gradually change into $2$ during updates. 

\begin{figure}[!h]
\vspace*{-2mm}
\centering
\centerline{
\begin{tabular}{cc}
\includegraphics[width=.18\textwidth,height=1.4in]{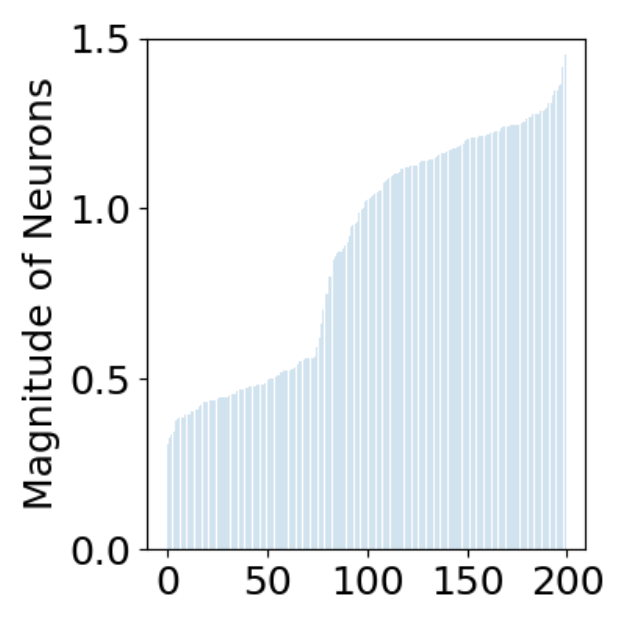}  
&
\hspace*{-1mm}
\includegraphics[width=.18\textwidth,height=1.4in]{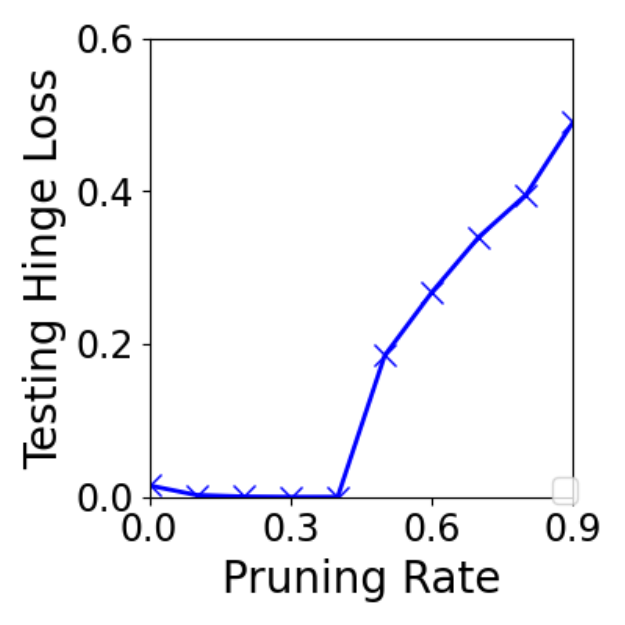}\vspace*{-1mm}
\\
(A) 
& \hspace*{-2mm} (B)
\end{tabular}}
\vspace*{-1mm}
\caption{(A) The magnitude of the trained $\bfW_O$ neurons (B) The testing performance of magnitude-based pruning with different pruning rate.}\label{fig: prune}
\vspace{-1mm}
\end{figure}

We next investigate the magnitude-based pruning on $\bfW_O$. Figure \ref{fig: prune} (A) shows that around an $80/200=0.4$ fraction of the trained $\bfW_O$ neurons have a small magnitude compared with the remaining ones, which first verifies (\ref{eqn: small WO}). Figure \ref{fig: prune} (B) illustrates that when the pruning rate is lower than $0.4$, i.e., pruning on these small neurons, the generalization performance does not become worse. When the pruning rate is larger than $0.4$, which means pruning neurons with a large magnitude, the testing error increases as the pruning rate increases. This finding justifies Corollary \ref{cor: pruning}.

\section{Conclusion}
This paper provides a theoretical analysis of the sparse and low-rank properties of trainable parameters and their gradient updates, respectively, by characterizing the direction and magnitude of a one-layer single-head trained Transformer. The conclusion indicates that the rank of gradient updates is determined by the number of label-relevant patterns. Our result also theoretically shows that pruning a constant fraction of neurons with a small magnitude has a trivial impact on the generalization, while pruning neurons with a large magnitude significantly harms the performance.

Future directions include analyzing the optimal condition of the LoRA algorithm and designing better methods for Transformer-based models via low-rank factorization and model pruning. We see no ethical or immediate negative societal consequence of our work.

\end{document}